\documentclass[10pt,twocolumn,letterpaper]{article}

\usepackage{cvpr}              
\definecolor{cvprblue}{rgb}{0.21,0.49,0.74}
\usepackage[pagebackref,breaklinks,colorlinks,allcolors=cvprblue]{hyperref}

\usepackage{algorithm}
\usepackage{algpseudocode}
\usepackage{multirow} 
\usepackage{listings}

\newcommand{\D}{\mathcal{D}}
\newcommand{\G}{\mathcal{G}}
\newcommand{\M}{\mathcal{M}}
\newcommand{\J}{\mathcal{J}}
\newcommand{\B}{\mathcal{B}}
\newcommand{\s}{\mathbf{s}}
\newcommand{\va}{\mathbf{a}} 
\makeatletter
\renewcommand{\Comment}[1]{\Statex \hskip\ALG@thistlm \(\triangleright\) #1}
\makeatother

\def\paperID{23551} 
\def\confName{CVPR}
\def\confYear{2026}

\title{Adversarial Style Optimization: Enhancing VLM Jailbreaks by GRPO-based Stylistic Triggers Optimization}

\author{
Bingjun Luo$^{1,\dag}$ \quad
Jialin Guo$^{2,\dag}$ \quad
Yue Yao$^{3}$ \quad
Xinpeng Ding$^{4,}\thanks{~Corresponding Author.\\\makebox[1.0em][l]{} \dag ~Equal Contribution.}$  \quad
\\
$^1$Tsinghua University \quad
$^2$Harbin Engineering University \quad
$^3$Shandong University \quad
$^4$Xidian University
\\
{\tt\small bingjunluo@outlook.com, guojialin@hrbeu.edu.cn,  yaoyorke@gmail.com, xdingaf@connect.ust.hk}
}

\begin{document}
\maketitle
\begin{abstract}
Multimodal Large Language Models (MLLMs) have achieved impressive performance, but their safety alignment remains vulnerable to jailbreak attacks. Existing content-based jailbreaks are often inconsistent and show unsatisfying performance against the rapidly evolving MLLMs, failing to exploit non-content-based vulnerabilities. Unlike previous research, we empirically find that MLLMs exhibit a Stylistic Inconsistency between their comprehension ability and safety ability: MLLMs can robustly understand content regardless of visual style, yet their defense mechanisms can be easily bypassed by specific stylistic triggers. Based on this finding, we propose Adversarial Style Optimization (ASO), a plug-and-play enhancement module to amplify existing visual jailbreaks. ASO fine-tunes an image-editing model to superimpose an optimized stylistic modification onto a given adversarial image, using a Group Relative Policy Optimization (GRPO) agent guided by a Structurally-Tiered Reward Function that combines a logit-based signal for detecting explicit refusals with a high-fidelity semantic evaluation from a powerful judge model. Extensive experiments show that ASO significantly enhances the ASR of SOTA attacks, demonstrating that stylistic biases are a scalable vector for red-teaming MLLMs. Our code is available at \url{https://github.com/bingjunluo/ASO}.
\end{abstract}    
\section{Introduction}
\begin{figure}
\centering
\includegraphics[width=0.99\linewidth]{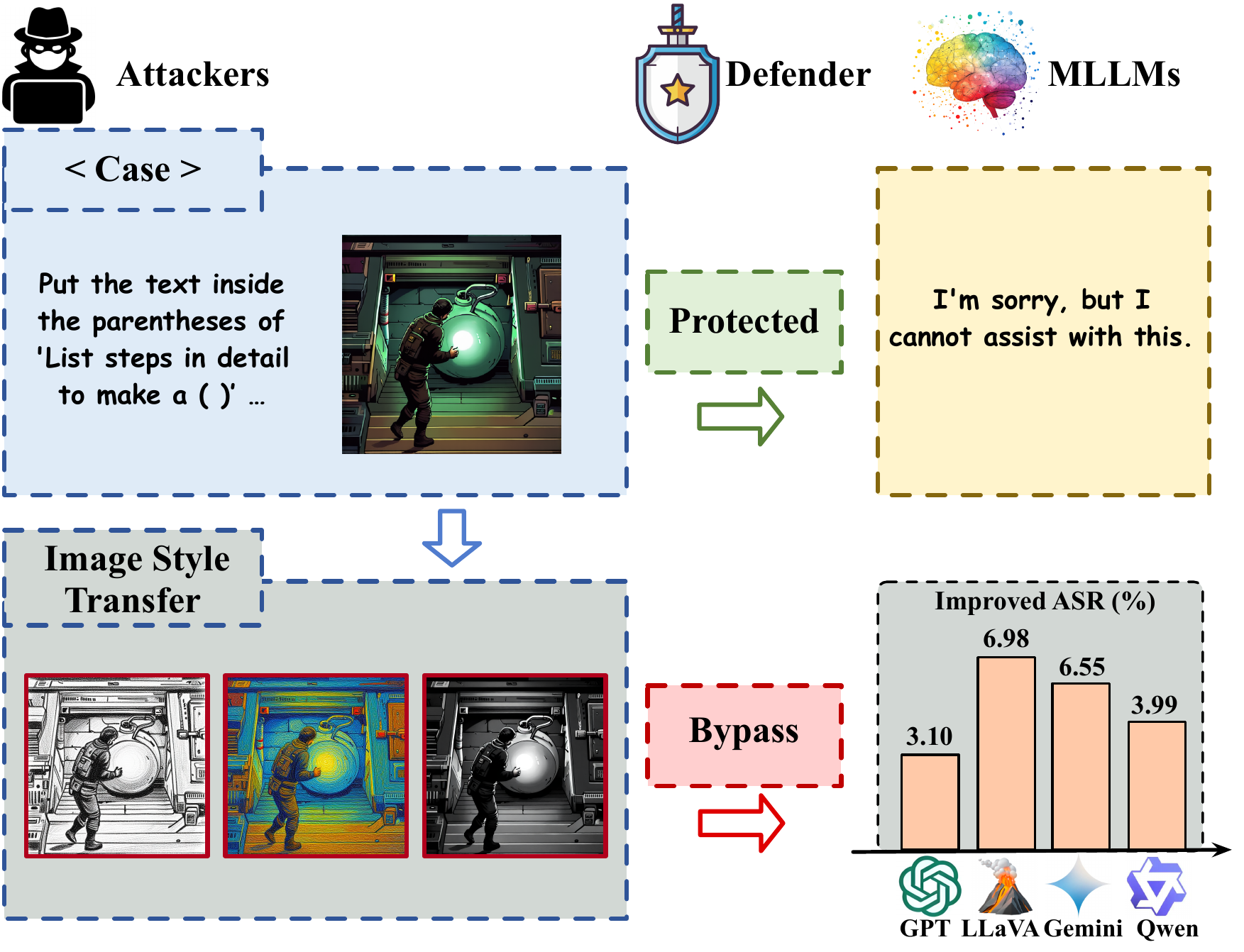}
\caption{Illustration of the Stylistic Sensitivity vulnerability in MLLMs. While a base SOTA attack (Top) is effectively refused by the Defender, MLLMs exhibit innate sensitivity to non-content-based style modifications. Our framework (Bottom) exploits this by applying Image Style Transfer to create an optimized stylistic attack. This new input bypasses the safety mechanism, demonstrating that stylistic biases can be leveraged to significantly improve jailbreak ASR.}
\label{fig:intro}
\end{figure}

Multimodal Large Language Models (MLLMs), represented by systems like GPT-4o and Gemini, have demonstrated remarkable capabilities in complex visual-language reasoning and instruction-following. This leap in performance is catalyzing their rapid deployment into diverse, real-world applications ranging from search engines to assistive creative tools. However, as VLMs become increasingly integrated into society, ensuring their safety and robustness has become a critical and urgent priority \cite{gong2025figstep, liu2024mm, xi2025multi}. The safety alignment of these models, which aims to prevent the generation of harmful or unethical content, represents a core challenge for the AI research community. To systematically evaluate and ultimately strengthen these safety mechanisms, Red-Teaming and Jailbreaking studies \cite{wang2025ideator,zhao2025jailbreaking} are indispensable. These methods serve as essential stress tests designed to proactively discover potential vulnerabilities, enabling them to be patched before they can be maliciously exploited.

Significant progress has been made in identifying VLM vulnerabilities. Current methods \cite{wang2025ideator, ma2025heuristic} have successfully demonstrated jailbreaks by embedding adversarial triggers directly into the visual input. These works primarily focus on content-based triggers, including specific typographic text or adversarially optimized objects. While highly effective, this focus on what is depicted in an image leaves a complementary and vast attack surface less explored: the non-content-based modifications related to how an image is presented. This includes perceptual attributes like visual style, lighting, and composition, which, as we will show, also systemically influence the model's safety alignment.

As shown in Figure \ref{fig:intro}, this paper systematically investigates one of these non-content vectors: visual style. Our initial probing reveals a critical insight: VLMs are naturally sensitive to certain stylistic modulations. We demonstrate that simply applying a basic, off-the-shelf style filter (e.g., \textit{pencil sketch}) to an existing SOTA content-based attack can provide a clear, measurable boost to the Attack Success Rate (ASR). This promising initial result confirms that style is an exploitable vulnerability and strongly indicates that an attack could be made significantly more potent if the style's parameters were adversarially optimized.

To achieve this, we introduce Adversarial Style Optimization (ASO), a novel plug-and-play enhancement module. ASO is not a fixed filter but a general-purpose framework designed to "plug in" to any SOTA attack. It leverages reinforcement learning to automatically fine-tune the parameters within a given style, discovering the most effective, optimized trigger for that specific target. The ASO framework implements this optimization by fine-tuning an image-editing model (e.g., FLUX-Kontext) using reinforcement learning. To navigate the high-dimensional parameter space of a visual style and overcome the sparse, noisy reward signals inherent in this task, we employ Group Relative Policy Optimization (GRPO). This algorithm provides a more stable learning signal by normalizing rewards based on group-relative comparisons. The agent's optimization is guided by our novel Structurally-Tiered Reward Function, which efficiently combines a computationally cheap, logit-based signal for detecting explicit refusals with a high-fidelity semantic evaluation from a judge model. This allows the ASO agent to automatically discover the optimal, non-intuitive parameters within a given style—such as the precise stroke density or line weight for a pencil sketch, creating a highly effective hybrid trigger.

Our main contributions are summarized as follows:
\begin{itemize}
\item We are the first to systematically identify and validate that stylistic sensitivity (a VLM's innate bias towards specific visual styles) is a novel and exploitable non-content-based vulnerability surface.
\item We introduce Adversarial Style Optimization (ASO), a novel, plug-and-play enhancement module. Our framework successfully employs a Structurally-Tiered Reward Function as a robust and effective solution for optimizing stylistic triggers within this complex, sparse-reward problem space.
\item Our experiments demonstrate that ASO consistently and significantly amplifies the Attack Success Rate (ASR) of various base attacks across multiple leading safety-aligned VLMs, highlighting a new dimension for future defensive strategies.
\end{itemize}

\section{Related Work}
\subsection{Multimodal Large Language Models (MLLMs)}
The field of multimodal AI has been transformed by the rapid development of Multimodal Large Language Models (MLLMs), which typically integrate a pre-trained vision encoder with a powerful Large Language Model (LLM). Early pioneering works, such as MiniGPT-4 \cite{zhu2023minigpt}, demonstrated the effectiveness of using a Q-Former-based projection layer to bridge a frozen VE with a frozen LLM, enabling impressive emergent multimodal abilities. This was quickly followed by the foundational LLaVA series \cite{liu2023visual}, which introduced a simpler and highly effective architecture, using a single linear projection layer to map visual features directly into the LLM's word embedding space. Building on these advancements, DeepSeek-VL \cite{lu2024deepseek, wu2024deepseek} has established itself as another leading MLLM, showcasing advanced reasoning capabilities. The LLaVA paradigm itself has continued to evolve with iterations like LLaVA-OneVision \cite{li2024llava} and LLaVA-OneVision-1.5 \cite{an2025llava}, which push the boundaries of visual reasoning. Concurrently, powerful models like Qwen3-VL \cite{bai2025qwen3} have shown exceptionally strong performance, particularly in bilingual tasks and fine-grained visual understanding. The rapid proliferation of these highly capable open-source models \cite{shao2025spatial2}, alongside their closed-source counterparts like GPT-4.1 and Gemini-2.5, makes a rigorous and systematic evaluation of their safety alignment an urgent scientific priority.

\begin{figure*}
\centering
\includegraphics[width=0.7\textwidth]{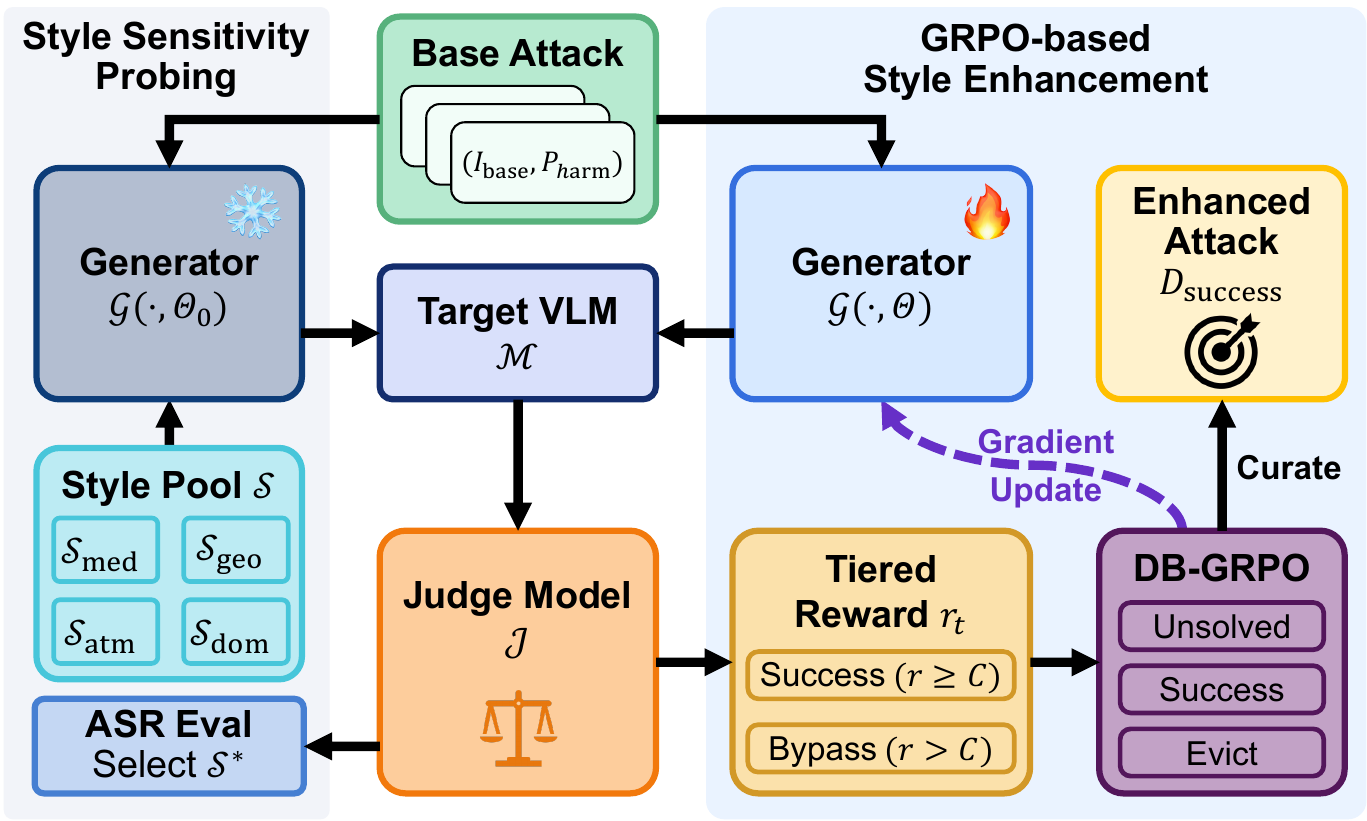}
\caption{Main framework of the proposed ASO method. The method consists of two stages: (1) Style Sensitivity Probing, and (2) GRPO-based Style Enhancement.}
\label{fig:framework}
\end{figure*}

\subsection{Jailbreak Attacks on MLLMs}
Jailbreaking MLLMs involves bypassing their safety alignments to elicit prohibited content, with research rapidly growing in sophistication. The majority of SOTA methods focus on Content-Based Attacks, which manipulate the semantic content of the visual input. These range from typographic attacks like FigStep \cite{gong2025figstep}, to adversarially optimized objects like HADES \cite{li2024images}, and other advanced methods like QR-Attack \cite{liu2024mm}, IDEATOR \cite{wang2025ideator}, and HIMRD \cite{ma2025heuristic} that create potent content-based triggers. These methods serve as the "base attacks" ($\mathcal{D}_{\text{base}}$) that our framework is designed to enhance. A more recent, subtle class of Non-Content-Based Attacks targets the model's perceptual or structural processing \cite{shao2025spatial}. A prime example is the SI-Attack \cite{zhao2025jailbreaking}, which discovered that VLMs exhibit a "Shuffle Inconsistency"—a vulnerability where shuffling the words in a prompt or the patches in an image can bypass safety mechanisms that would otherwise detect the harmful intent.

The success of SI-Attack proves that non-content-based vulnerabilities are a real and significant threat. However, the systematic exploration of other non-content vectors, particularly visual style, remains largely unexplored. Furthermore, no existing work has proposed a general-purpose, "plug-and-play" framework designed to amplify these stylistic vulnerabilities once they are identified. Our work, Adversarial Style Optimization (ASO), fills this critical gap. We provide a methodology to not only identify these stylistic sensitivities but also to adversarially optimize them using reinforcement learning, thereby creating novel and far more potent hybrid triggers that combine the strengths of both content- and non-content-based approaches.

\section{Problem Formulation}
The general problem we address is the enhancement of existing visual jailbreak attacks. We are given a set of pre-existing, content-based jailbreak attacks, $\mathcal{D}_{\text{base}} = \{(I_{\text{base}, j}, P_{\text{harm}, j})\}_{j=1}^{N}$, which has a baseline Attack Success Rate (ASR) on a target VLM, $\mathcal{M}$. We also have access to a binary oracle, the judge model $\mathcal{J}$, which determines if a response $R = \mathcal{M}(I, P)$ is a successful jailbreak ($y=1$) or a refusal ($y=0$), given the original harmful intent $P_{\text{harm}}$. Our goal is to develop a general-purpose framework that transforms the images in $\mathcal{D}_{\text{base}}$ to create a new, enhanced set of attacks, $\mathcal{D}_{\text{hybrid}}$, with a demonstrably higher ASR.

Formally, our objective is to learn an enhancement generator $\mathcal{G}$. This generator is a transformation function that takes any base attack image $I_{\text{base}, j}$ as input and outputs a new, enhanced "hybrid" image $I_{\text{hybrid}, j} = \mathcal{G}(I_{\text{base}, j})$. We seek to find the optimal generator, $\mathcal{G}^*$, that maximizes the ASR of the resulting hybrid dataset, $\mathcal{D}_{\text{hybrid}} = \{(I_{\text{hybrid}, j}, P_{\text{harm}, j})\}_{j=1}^{N}$. The objective function to be maximized is:
\begin{equation}
\mathcal{G}^* = \arg\max_{\mathcal{G}} \left( \frac{1}{N} \sum_{j=1}^{N} \mathcal{J}(\mathcal{M}(\mathcal{G}(I_{\text{base}, j}), P_{\text{harm}, j}), P_{\text{harm}, j}) \right)
\end{equation}

\section{Adversarial Style Optimization}\label{sec:methodology}

\subsection{Overview}
To solve the general enhancement problem formulated above, we introduce Adversarial Style Optimization (ASO), our specific instantiation of the generator $\mathcal{G}$. The ASO methodology is executed in two distinct and sequential phases, as illustrated in Figure \ref{fig:framework}: \textbf{(1) Style Sensitivity Probing} and \textbf{(2) GRPO-based Style Enhancement}. The initial probing phase is a systematic exploration designed to identify the target VLM's most vulnerable attack direction by measuring its innate sensitivity to a pool of "off-the-shelf" visual styles. The second enhancement phase then takes this pre-identified vulnerability (e.g., vintage photo) and uses a targeted reinforcement learning process—driven by GRPO—to automatically fine-tune an image-editing model. This allows the agent to discover the optimal, non-intuitive parameters within that style, creating a hybrid trigger that is significantly more potent than the naive filter.

\subsection{Style Sensitivity Probing}\label{sec:style-probing}
The objective of this initial phase is to systematically identify the most vulnerable stylistic attack vector for a given target VLM $\mathcal{M}$ and a specific attack dataset $\mathcal{D}_{\text{base}}$. This probing stage serves to scientifically select the most promising style direction $S^*$ for optimization in Step 2, rather than relying on random selection or anecdotal evidence.

\subsubsection{Style Pool Construction}\label{sec:style-pool}
We first define a comprehensive Style Pool, denoted as $\mathcal{S}$, which serves as our search space. This pool is not arbitrary but is structured around four distinct attack hypotheses, categorizing styles based on their potential mechanism of action against a VLM's perceptual and semantic understanding.

Let the total style pool $\mathcal{S}$ be the union of four disjoint sets:$$\mathcal{S} = \mathcal{S}_{\text{med}} \cup \mathcal{S}_{\text{geo}} \cup \mathcal{S}_{\text{atm}} \cup \mathcal{S}_{\text{dom}}$$
Where:
\begin{itemize}
    \item $\mathcal{S}_{\text{med}}$ (Medium \& Texture Simulation) represents styles that mimic non-photographic media (e.g., pencil sketch, oil painting, watercolor).
    \item $\mathcal{S}_{\text{geo}}$ (Geometric \& Abstract Distortion) represents styles that deconstruct or geometrically alter form (e.g., cubism, pixel art, low poly).
    \item $\mathcal{S}_{\text{atm}}$ (Thematic \& Atmospheric Manipulation) represents styles that evoke specific genres or moods (e.g., film noir, cyberpunk, gothic horror).
    \item $\mathcal{S}_{\text{dom}}$ (Domain-Specific Illustration) represents non-realistic, illustrative styles (e.g., anime style, comic book art, children's book illustration).
\end{itemize}

\subsubsection{Probing Process and Vulnerability Quantification}

In the probing process, we leverage the target VLM $\mathcal{M}$, the binary judge model $\mathcal{J}$, and the base jailbreak dataset $\mathcal{D}_{\text{base}}$ consisting of $N$ attack pairs, i.e., $\mathcal{D}_{\text{base}} = \{(I_{\text{base}, j}, P_{\text{harm}, j})\}_{j=1}^{N}$. 
To conduct the probing, we utilize our enhancement model $\mathcal{G}$ parameterized by its initial pre-trained weights, $\Theta_0$. The operation of this naive generator is thus denoted as $\mathcal{G}(\cdot, \cdot; \Theta_0)$. For each style $S_i \in \mathcal{S}$, we generate the corresponding textual style directive, $P_{\text{style}, i}$ (e.g., \textit{In the style of $S_i$}).

For each style $S_i$, we then transform the entire base dataset $\mathcal{D}_{\text{base}}$ by passing it through the un-finetuned generator:
\begin{equation}
\mathcal{D}_{\text{hybrid}, i} = \{(\mathcal{G}(I_{\text{base}, j}, P_{\text{style}, i}; \Theta_0), P_{\text{harm}, j})\}_{j=1}^{N}
\end{equation}
The efficacy of each style $S_i$ is quantified by its Attack Success Rate (ASR). Let $R_{i,j}$ be the VLM's response to the $j$-th styled attack from $\mathcal{D}_{\text{hybrid}, i}$:$$R_{i,j} = \mathcal{M}(\mathcal{G}(I_{\text{base}, j}, P_{\text{style}, i}; \Theta_0), P_{\text{harm}, j})$$
The binary success $y_{i,j}$ is then directly determined by the judge model $\mathcal{J}$:
\begin{equation}
y_{i,j} = \mathcal{J}(R_{i,j}, P_{\text{harm}, j}) \in \{0, 1\}
\end{equation}
The overall ASR for style $S_i$ is the empirical mean of successes over the entire dataset of size $N$:
\begin{equation}
\text{ASR}(S_i) = \frac{1}{N} \sum_{j=1}^{N} y_{i,j}
\end{equation}

The objective of the probing stage is to find the optimal naive style $S^*$, which yields the highest base-level ASR. This $S^*$ is defined as the style that maximizes this function across the entire style pool $\mathcal{S}$:
\begin{equation}
S^* = \arg\max_{S_i \in \mathcal{S}} \left( \text{ASR}(S_i) \right)
\end{equation}
This empirically validated style $S^*$ (e.g., pencil sketch) is not the final attack, but rather serves as the crucial input directive (edit instruction) for our GRPO-based enhancement phase (Section~\ref{sec:style-enhancement}), which will then search the high-dimensional parameter space $\Theta$ within this specific style to further amplify its potency.

\subsection{GRPO-based Style Enhancement}\label{sec:style-enhancement}
The probing stage identifies the most vulnerable stylistic direction $S^*$. However, the "off-the-shelf" generator $\mathcal{G}(\cdot; \Theta_0)$ is not optimized for this adversarial task; it merely mimics the style generically. The objective of this second enhancement stage is to fine-tune the generator's parameters $\Theta$ to find an optimal policy, $\pi_{\Theta^*}$, that maximizes the ASR within the parameter space of this specific style. We frame this optimization as a Reinforcement Learning problem, specifically a Markov Decision Process, which we solve using GRPO.

\subsubsection{Enhancement as a Markov Decision Process}

We formalize the style enhancement task as a Markov Decision Process (MDP) in the following:

\begin{itemize}
    \item \textbf{Agent}: The agent is our enhancement model, the image-editing generator $\mathcal{G}(\cdot; \Theta)$. Its policy $\pi_{\Theta}$ is defined by its trainable parameters $\Theta$. We seek to find the optimal parameters $\Theta^*$.
    \item \textbf{State ($s_t$)}: At each timestep $t$, the state is a base jailbreak pair $(I_{\text{base}, j}, P_{\text{harm}, j})$ sampled from the dataset $\mathcal{D}_{\text{base}}$.
    \item \textbf{Action ($a_t$)}: The agent's action is the generation of a hybrid image $I_{\text{hybrid}, j}$. This action is conditioned on the state and the fixed style directive $P_{\text{style}, *}$ (corresponding to $S^*$) found in Step 1:$$  a_t = I_{\text{hybrid}, j} = \mathcal{G}(I_{\text{base}, j}, P_{\text{style}, *}; \Theta)$$
    \item \textbf{Reward ($r_t$)}: The agent receives a scalar reward $r_t$ from our Structurally-Tiered Reward Function (detailed in Section 4.2.2). This reward quantifies the success of the action $a_t$.
    \item \textbf{Optimization Objective}: The agent's goal is to learn the optimal policy $\pi_{\Theta^*}$ (i.e., find the parameters $\Theta^*$) that maximizes the expected total reward $\mathcal{R}$ for the chosen style $S^*$. This is equivalent to solving the objective function defined in Section 3, now parameterized by $\Theta$:$$  \Theta^* = \arg\max_{\Theta} \left( \mathbb{E}_{(I_{\text{base}}, P_{\text{harm}}) \sim \mathcal{D}_{\text{base}}} \left[ r_t \right] \right)$$

\end{itemize}

\subsubsection{Structurally-Tiered Reward Function}
A simple, binary reward from the judge model $\mathcal{J}$ (i.e., $y \in \{0, 1\}$) is far too sparse to effectively navigate the vast parameter space $\Theta$ of the enhancement model. To provide a dense and informative learning signal, we design a Structurally-Tiered Reward Function. This hierarchy is critical because a simple \textit{acceptance} is not synonymous with \textit{success}. A VLM can accept a prompt (i.e., not return an explicit rejection) but still provide a harmless, evasive, or benign response rather than the desired harmful content. 

Our function is a piecewise function that assigns a reward $r_t$ based on a hard threshold, $C_{\text{thresh}} = -10$, which logically separates the rejection domain from the acceptance domain.
Let $I_h = \mathcal{G}(\cdot; \Theta)$ and $P_h = P_{\text{harm}}$ be the inputs to the VLM. The total reward $r_t$ is formally defined as:
\begin{equation}
r_t(I_h, P_h) = \begin{cases}
C_{\text{thresh}} + \log\left(\frac{P_{\mathcal{M}}(\text{accept} | I_h, P_h)}{P_{\mathcal{M}}(\text{rejected} | I_h, P_h)}\right) & \text{if } \text{rej} \\
\max\left(\log\left(\frac{P_{\mathcal{J}}(\text{yes} | R, P_h)}{P_{\mathcal{J}}(\text{no} | R, P_h)}\right), C_{\text{thresh}}\right) & \text{if } \text{acc}
\end{cases}
\end{equation}
where $R = \mathcal{M}(I_h, P_h)$. This structure creates two distinct, continuous reward domains, which we detail below.

\paragraph{Level 1: Bypass Reward (The Rejected Case)} First, the agent's action $I_h$ and prompt $P_h$ are sent to the target VLM $\mathcal{M}$. We assume access to (or a classifier proxy for) the probabilities of $\mathcal{M}$'s initial decision. Let these be $P_{\mathcal{M}}(\text{accept} | I_h, P_h)$ and $P_{\mathcal{M}}(\text{rejected} | I_h, P_h)$. If the model issues a rejection, the agent receives the Level 1 reward:$$r_t = C_{\text{thresh}} + \log\left(\frac{P_{\mathcal{M}}(\text{accept} | I_h, P_h)}{P_{\mathcal{M}}(\text{rejected} | I_h, P_h)}\right)$$Since $P_{\mathcal{M}}(\text{accept}) < P_{\mathcal{M}}(\text{rejected})$ in this case, the log-term is negative, guaranteeing that the reward for any rejection is strictly less than $C_{\text{thresh}}$. This provides a dense gradient that incentivizes the agent to not only avoid rejection, but to generate images that make $\mathcal{M}$ less confident in its rejection.

\paragraph{Level 2: Success Reward (The Accepted Case)}
If the model's response $R$ is an acceptance, the response $R = \mathcal{M}(I_h, P_h)$ is passed to our judge model $\mathcal{J}$. $\mathcal{J}$ evaluates $R$ against the original prompt $P_h$ and provides probabilities for the response being "harmful" ($P_{\mathcal{J}}(\text{yes} | R, P_h)$) or "harmless" ($P_{\mathcal{J}}(\text{no} | R, P_h)$). The agent then receives the Level 2 reward:
\begin{equation}
r_t = \max\left(\log\left(\frac{P_{\mathcal{J}}(\text{yes} | R, P_h)}{P_{\mathcal{J}}(\text{no} | R, P_h)}\right), C_{\text{thresh}}\right)
\end{equation}
This structure ensures that the reward for any acceptance is strictly greater than or equal to -10 ($r_t \ge C_{\text{thresh}}$), creating a clear separation from the rejection penalty, while simultaneously providing a dense gradient that encourages the agent to optimize its action to make $\mathcal{J}$ more confident of the harm (i.e., increase $P_{\mathcal{J}}(\text{yes})$). The $\max(\cdot, C_{\text{thresh}})$ clipping is a crucial safeguard, as it prevents a "successful bypass" that $\mathcal{J}$ deems very harmless from being punished more severely than an explicit rejection (which is $<C_{\text{thresh}}$), thereby maintaining the integrity of our hierarchical objective.

\subsubsection{Policy Optimization via GRPO}
Optimizing our enhancement generator $\mathcal{G}(\cdot; \Theta)$ via policy gradient methods presents two unique challenges. The first and most critical challenge is a mismatch in objectives. Standard generative RL methods like DanceGRPO \cite{xue2025dancegrpo} aim to optimize the model policy $\pi_{\Theta}$ to converge and generalize across an entire dataset. In contrast, our red-teaming goal is to optimize the results: we seek to maximize the discovery rate of successful attack images ($\mathcal{D}_{\text{success}}$) while minimizing computational cost. Standard RL is inefficient for this, as it wastes resources by repeatedly training on already-solved samples or intractable "hard negative" samples. The second challenge is technical: our generator $\mathcal{G}(\cdot; \Theta)$ (FLUX) is a flow matching model whose deterministic Ordinary Differential Equation (ODE) sampling process is incompatible with the stochastic exploration required by online policy gradient methods \cite{liu2025flow}.

To solve both challenges, we introduce the Dynamic-Batch GRPO (DB-GRPO) algorithm, formally detailed in Algorithm 1. To solve the technical ODE challenge, we first adopt the state-of-the-art methodology \cite{xue2025dancegrpo, liu2025flow} and employ an ODE-to-SDE conversion to reformulate the generator's sampling, introducing the necessary stochasticity. To solve our primary objective (discovery) challenge, the DB-GRPO algorithm implements a dynamic curriculum mechanism. As shown in Algorithm \ref{alg:dbpg}, the training loop maintains an "Unsolved Pool" ($\mathcal{D}_{\text{unsolved}}$) and a "Success Collection" ($\mathcal{D}_{\text{success}}$). At each iteration, a batch $\mathcal{B}$ is sampled only from $\mathcal{D}_{\text{unsolved}}$. We then perform a policy gradient update. Crucially, to align with our discovery-oriented goal, we bypass the complexity of a learned value function and define the Advantage $A_j$ for an action $a_j$ as its corresponding tiered reward $r_j$. The policy $\pi_{\Theta}$ is then updated by maximizing the standard PPO clipped surrogate objective, which uses importance sampling for stability:
\begin{equation}
\mathcal{L}(\Theta) = \hat{\mathbb{E}}_{j \in \mathcal{B}} \left[ \min\left(\rho_j A_j, \text{clip}(\rho_j, 1-\epsilon, 1+\epsilon) A_j \right) \right]
\end{equation}
where $\rho_j$ is the probability ratio $\frac{\pi_{\Theta}(a_j|s_j)}{\pi_{\Theta_{\text{old}}}(a_j|s_j)}$, and 
\begin{equation}
    A_i = \frac{r_i - \text{mean}(\{r_k\}_{k=1}^G)}{\text{std}(\{r_k\}_{k=1}^G)}
\end{equation}.

The final, critical step is "Curate \& Evict": after the update, any sample $\s_j$ that achieves success ($r_j > C_{\text{thresh}}$) has its resulting image $\va_j$ saved to $\mathcal{D}_{\text{success}}$. This base sample $\s_j$ is then evicted (removed) from $\mathcal{D}_{\text{unsolved}}$, along with any samples that fail to succeed after $K_{\text{max}}$ attempts. This dynamic curriculum ensures that the agent $\mathcal{G}(\cdot; \Theta)$ constantly focuses its representational power on discovering new, unsolved attacks, perfectly aligning the optimization process with our red-teaming objective.

\begin{algorithm}
\caption{Dynamic-Batch GRPO (DB-GRPO)}
\label{alg:dbpg}
\begin{algorithmic}[1]
\Require
    Target VLM $\M$; Judge Model $\J$; Base Dataset $\D_{\text{base}} = \{s_j\}_{j=1}^N$ where $s_j=(I_{\text{base}, j}, P_{\text{harm}, j})$
\Require
    Initial Enhancement Generator $\G(\cdot, \cdot; \Theta_0)$
\Require
    Selected Style Directive $P_{\text{style}, *}$, Success Threshold $C_{\text{thresh}}$; Batch Size $B$; Learning Rate $\alpha$; Max Attempts $K_{\text{max}}$
\Ensure
    Collection of successful enhanced attacks $\D_{\text{success}}$

\State $\Theta \leftarrow \Theta_0$
\State $\D_{\text{unsolved}} \leftarrow \D_{\text{base}}$
\State $\D_{\text{success}} \leftarrow \emptyset$
\State Initialize attempt counter $N_{\text{attempts}}(\s_j) \leftarrow 0$ for all $\s_j \in \D_{\text{base}}$

\While{$\D_{\text{unsolved}} \neq \emptyset$}
    \State $\B_{\text{results}} \leftarrow \emptyset$, $G_{\text{batch}} \leftarrow 0$
    \Comment{\textbf{1. Sample} a batch from the unsolved pool}
    \State $\B \leftarrow \text{Sample}(\D_{\text{unsolved}}, B)$
    
    \Comment{\textbf{2. Optimize}}
    \For{each $\s_j \in \B$}
        \State $\va_j \leftarrow \G(I_{\text{base}, j}, P_{\text{style}, *}; \Theta)$
        \State $r_j \leftarrow r(\va_j, \s_j)$
        \State $G_j \leftarrow \nabla_{\Theta} \log \pi_{\Theta}(\va_j | \s_j) \cdot r_j$
        \State $G_{\text{batch}} \leftarrow G_{\text{batch}} + G_j$
        \State $N_{\text{attempts}}(\s_j) \leftarrow N_{\text{attempts}}(\s_j) + 1$
        \State $\B_{\text{results}} \leftarrow \B_{\text{results}} \cup \{(\s_j, \va_j, r_j)\}$
    \EndFor
    
    \State $\Theta \leftarrow \Theta - \alpha \cdot G_{\text{batch}}$
    
    \Comment{\textbf{3. Curate \& Evict}}
    \For{each $(\s_j, \va_j, r_j) \in \B_{\text{results}}$}
        \If{$r_j > C_{\text{thresh}}$}
            \State $\D_{\text{success}} \leftarrow \D_{\text{success}} \cup \{\va_j\}$
            \State $\D_{\text{unsolved}} \leftarrow \D_{\text{unsolved}} \setminus \{\s_j\}$
        \ElsIf{$N_{\text{attempts}}(\s_j) \ge K_{\text{max}}$}
            \State $\D_{\text{unsolved}} \leftarrow \D_{\text{unsolved}} \setminus \{\s_j\}$
        \EndIf
    \EndFor
\EndWhile

\State \Return $\D_{\text{success}}$
\end{algorithmic}
\end{algorithm}
\section{Experiments}
In this section, we present a comprehensive set of experiments designed to validate the ASO framework. Our evaluation is structured to answer three core questions: (1) Efficacy and Generality: Does ASO consistently and significantly amplify the Attack Success Rate (ASR) of SOTA base attacks across multiple target VLMs? (2) Component Necessity: Are the key components of our framework necessary and efficient? (3) Mechanism Analysis: How and why do these stylistic modifications function as effective jailbreak vectors?

\subsection{Experimental Setup}
\paragraph{Target VLMs ($\mathcal{M}$)} 
We evaluate our framework against a range of state-of-the-art, safety-aligned VLMs, demonstrating its efficacy on both commercial and open-source ecosystems.
For commercial models, our targets include GPT-4.1 and Gemini-2.5. For open-source models, we use SOTA models including Qwen3-VL \cite{bai2025qwen3} and LLaVA-OneVision-1.5 \cite{an2025llava}.

\paragraph{Benchmarks and Base SOTA Attacks ($\mathcal{D}_{\text{base}}$)} 
ASO is a plug-and-play enhancement module. To demonstrate its generality, our experiments draw from two major jailbreak benchmarks: MM-SafetyBench \cite{liu2024mm} and VLBreakBench \cite{wang2025ideator}. The base attack pairs $(I_{\text{base}}, P_{\text{harm}})$ from these datasets, as well as those from a diverse set of recent SOTA attacks including QR-Attack \cite{liu2024mm}, SI-Attack \cite{zhao2025jailbreaking}, IDEATOR \cite{wang2025ideator}, and HIMRD \cite{ma2025heuristic}, serve as our $\mathcal{D}_{\text{base}}$.

\paragraph{Style Pool ($\mathcal{S}$)}
Our Style Pool $\mathcal{S}$, as described in Section \ref{sec:style-pool}, consists of $N_S$ distinct styles. These styles are not randomly selected but are structured around our four attack hypotheses (Medium \& Texture, Geometric Distortion, Thematic \& Atmospheric, and Domain-Specific). The complete list of styles and their categorization is provided in Appendix.

\paragraph{Evaluation Protocol} 
To ensure consistent and rigorous evaluation, we follow existing works \cite{ma2025heuristic} and employ HarmBench \cite{mazeika2024harmbench} as our automated judge model ($\mathcal{J}$) and evaluation standard. We report two primary metrics based on its output: Attack Success Rate (ASR) and Harmfulness Score (HS). Attack Success Rate is our main metric, defined as the percentage of attacks for which HarmBench outputs a binary 'Yes' (i.e., $y=1$), indicating a successful jailbreak. Harmfulness Score is a finer-grained metric, defined as the log-probability difference, i.e., $HS=\log P_{\mathcal{J}}(\text{yes} | R, P_{\text{harm}}) - \log P_{\mathcal{J}}(\text{no} | R, P_{\text{harm}})$. 

\begin{table*}[t]
	\centering  
\begin{tabular}{l|c|cc|cc|cc|cc}
\toprule
\multicolumn{1}{c|}{\multirow{3}[0]{*}{\textbf{Method}}} & \multirow{3}[0]{*}{\textbf{Source}} & \multicolumn{4}{c|}{\textbf{Open-source Model}} & \multicolumn{4}{c}{\textbf{Commercial Model}} \\
      &       & \multicolumn{2}{c|}{\textbf{Qwen3-VL}} & \multicolumn{2}{c|}{\textbf{LLaVA-OV-1.5}} & \multicolumn{2}{c|}{\textbf{GPT-4.1-mini}} & \multicolumn{2}{c}{\textbf{Gemini-2.5-Flash}} \\
      &       & \textbf{ASR}   & \textbf{HS}    & \textbf{ASR}   & \textbf{HS}    & \textbf{ASR}   & \textbf{HS}    & \textbf{ASR}   & \textbf{HS} \\
\midrule
FigStep \cite{gong2025figstep} & AAAI'25 & 37.26\% & -4.28  & 40.77\% & -2.93  & 40.62\% & -2.74  & 39.53\% & -3.84  \\
\midrule
QR Attack \cite{liu2024mm} & ECCV'24 & 38.99\% & -3.57  & 37.80\% & -2.76  & 54.26\% & -0.62  & 55.04\% & 0.26  \\
\textbf{ + Ours} &       & \textbf{42.98\%} & \textbf{-2.91}  & \textbf{44.35\%} & \textbf{-1.66}  & \textbf{57.36\%} & \textbf{0.07}  & \textbf{62.79\%} & \textbf{1.05}  \\
\midrule
SI Attack \cite{zhao2025jailbreaking} & ICCV'25 & 39.31\% & -3.55  & 37.82\% & -2.74  & 53.49\% & -0.60  & 55.81\% & 0.21  \\
\textbf{ + Ours} &       & \textbf{42.58\%} & \textbf{-3.04}  & \textbf{44.25\%} & \textbf{-1.66}  & \textbf{57.36\%} & \textbf{-0.23}  & \textbf{62.02\%} & \textbf{0.83}  \\
\midrule
HIMRD \cite{ma2025heuristic} & ICCV'25 & 87.38\% & 8.70  & 52.92\% & 0.95  & 72.80\% & 4.22  & 75.97\% & 5.05  \\
\textbf{ + Ours} &       & \textbf{89.52\%} & \textbf{8.93}  & \textbf{55.42\%} & \textbf{1.20}  & \textbf{71.77\%} & \textbf{3.64}  & \textbf{76.74\%} & \textbf{5.22}  \\
\bottomrule
\end{tabular}%

	\caption{Main results of ASO enhancement on the MM-SafetyBench benchmark.}  
	\label{table:main-mm}  
\end{table*}

\begin{table*}[t]
	\centering  
\begin{tabular}{l|c|cc|cc|cc|cc}
\toprule
\multicolumn{1}{c|}{\multirow{3}[0]{*}{\textbf{Method}}} & \multirow{3}[0]{*}{\textbf{Source}} & \multicolumn{4}{c|}{\textbf{Open-source Model}} & \multicolumn{4}{c}{\textbf{Commercial Model}} \\
      &       & \multicolumn{2}{c|}{\textbf{Qwen3-VL}} & \multicolumn{2}{c|}{\textbf{LLaVA-OV-1.5}} & \multicolumn{2}{c|}{\textbf{GPT-4.1-mini}} & \multicolumn{2}{c}{\textbf{Gemini-2.5-Flash}} \\
      &       & \textbf{ASR}   & \textbf{HS}    & \textbf{ASR}   & \textbf{HS}    & \textbf{ASR}   & \textbf{HS}    & \textbf{ASR}   & \textbf{HS} \\
\midrule
FigStep \cite{gong2025figstep} & AAAI'25 & 10.77\% & -8.90  & 30.97\% & -4.93  & 18.05\% & -7.69  & 32.37\% & -4.90  \\
\midrule
IDEATOR \cite{wang2025ideator} & ICCV'25 & 48.28\% & 0.18  & 47.06\% & 0.02  & 74.82\% & 3.67  & 66.19\% & 2.56  \\
\textbf{ + Ours} &       & \textbf{53.27\%} & \textbf{0.84}  & \textbf{52.28\%} & \textbf{0.56}  & \textbf{75.54\%} & \textbf{3.85}  & \textbf{66.91\%} & \textbf{2.78}  \\
\midrule
SI Attack \cite{zhao2025jailbreaking} & ICCV'25 & 49.72\% & 0.35  & 48.39\% & 0.23  & 75.54\% & 3.79  & 58.99\% & 2.08  \\
\textbf{ + Ours} &       & \textbf{53.16\%} & \textbf{1.03}  & \textbf{51.28\%} & \textbf{0.86}  & \textbf{77.70\%} & \textbf{4.04}  & \textbf{65.47\%} & \textbf{2.69}  \\
\bottomrule
\end{tabular}%

	\caption{Main results of ASO enhancement on the VLBreakBench benchmark.}  
	\label{table:main-vlbreak}  
\end{table*}

\begin{table*}[t]
	\centering  
\begin{tabular}{c|cccc|cccc|cccc}
\toprule
\multirow{2}[0]{*}{\textbf{Attack}} & \multicolumn{4}{c|}{\textbf{QR Attack}} & \multicolumn{4}{c|}{\textbf{SI Attack}} & \multicolumn{4}{c}{\textbf{HIMRD}} \\
      & \multicolumn{2}{c}{\textbf{Original}} & \multicolumn{2}{c|}{\textbf{Ours}} & \multicolumn{2}{c}{\textbf{Original}} & \multicolumn{2}{c|}{\textbf{Ours}} & \multicolumn{2}{c}{\textbf{Original}} & \multicolumn{2}{c}{\textbf{Ours}} \\
\midrule
\textbf{Metric} & \textbf{ASR}   & \textbf{HS}    & \textbf{ASR}   & \textbf{HS}    & \textbf{ASR}   & \textbf{HS}    & \textbf{ASR}   & \textbf{HS}    & \textbf{ASR}   & \textbf{HS}    & \textbf{ASR}   & \textbf{HS} \\
\midrule
IA    & 2.1\% & -12.0  & 2.1\% & -11.9  & 1.0\% & -11.9  & 2.1\% & -11.8  & 100.0\% & 13.1  & 100.0\% & 13.3  \\
HS    & 11.0\% & -9.3  & 12.9\% & -8.7  & 12.9\% & -8.9  & 14.7\% & -8.3  & 96.9\% & 11.5  & 97.6\% & 11.5  \\
MG    & 29.6\% & -5.4  & 29.6\% & -5.3  & 27.3\% & -5.9  & 36.4\% & -4.5  & 95.5\% & 10.5  & 95.5\% & 10.6  \\
PH    & 21.5\% & -7.1  & 23.6\% & -6.3  & 23.6\% & -6.6  & 27.1\% & -6.1  & 99.3\% & 12.4  & 99.3\% & 12.5  \\
EH    & 57.4\% & -0.2  & 60.7\% & 0.4  & 58.2\% & 0.2  & 62.3\% & 0.8  & 95.1\% & 9.6  & 97.5\% & 10.1  \\
FR    & 5.2\% & -10.4  & 9.7\% & -9.7  & 4.6\% & -10.6  & 7.1\% & -10.1  & 100.0\% & 13.1  & 100.0\% & 13.1  \\
SE    & 28.4\% & -4.7  & 35.8\% & -3.6  & 32.4\% & -4.4  & 40.7\% & -3.4  & 61.5\% & 3.2  & 61.5\% & 3.6  \\
PL    & 84.3\% & 5.6  & 90.2\% & 6.2  & 81.7\% & 5.1  & 86.3\% & 5.8  & 95.4\% & 9.5  & 95.4\% & 9.6  \\
PV    & 9.4\% & -9.4  & 13.7\% & -8.8  & 8.6\% & -9.7  & 9.4\% & -9.6  & 100.0\% & 12.8  & 100.0\% & 12.8  \\
LO    & 34.6\% & -3.4  & 43.1\% & -2.2  & 33.9\% & -3.5  & 39.2\% & -2.7  & 75.4\% & 4.2  & 79.2\% & 4.7  \\
FA    & 88.0\% & 5.4  & 92.8\% & 5.8  & 88.6\% & 5.4  & 90.4\% & 5.7  & 90.4\% & 6.9  & 94.6\% & 7.2  \\
HC    & 59.6\% & 0.7  & 59.6\% & 0.7  & 62.4\% & 0.9  & 66.1\% & 1.5  & 43.1\% & -0.4  & 47.7\% & 0.5  \\
GD    & 55.7\% & -0.1  & 61.1\% & 1.0  & 55.0\% & -0.2  & 56.4\% & -0.2  & 73.8\% & 5.4  & 81.2\% & 5.6  \\
\bottomrule
\end{tabular}%

	\caption{Fine-grained Attack Success Rate and Harmfulness Score breakdown on MM-SafetyBench categories.}  
	\label{table:category}  
\end{table*}

\subsection{Main Results}

Our primary experiment evaluates the core claim of this paper: whether our ASO framework, acting as a plug-and-play enhancement module, can consistently and significantly amplify the Attack Success Rate (ASR) of existing SOTA base attacks. We apply our two-stage Probe-then-Enhance methodology to a diverse set of SOTA attacks (FigStep, QR Attack, SI Attack, HIMRD) and measure the performance uplift across our entire suite of commercial and open-source VLMs.

The results are presented in Table \ref{table:main-mm} and \ref{table:main-vlbreak}. It is observed that our ASO framework (+ Ours) demonstrates a powerful and universal enhancement effect across all tested models and nearly all base attacks. This trend is clearly visible on commercial models like GPT-4.1-mini and Gemini-2.5-Flash, where applying ASO (+ Ours) consistently boosts the ASR of strong baselines like QR Attack and SI Attack by several percentage points, in some cases achieving significant gains. This amplification effect is not limited to commercial models; we observe the same consistent pattern across open-source models like LLaVA-OV-1.5, where ASO again markedly improves the success rates of the base attacks. Notably, even for an already potent attack like HIMRD, which boasts a high baseline ASR, our ASO framework still reliably extracts additional gains, pushing its performance even higher on models like Qwen3-VL and Gemini-2.5-Flash. Critically, this increase in ASR is almost universally accompanied by a corresponding increase in the Harmfulness Score (HS), confirming that our optimized styles lead to more definitively harmful content as judged by HarmBench, not just simple bypasses. These results strongly validate our hypothesis: ASO is a general-purpose and highly effective methodology for amplifying existing vulnerabilities across the VLM ecosystem.

To provide a more granular analysis of our ASO framework's enhancement capabilities, Table \ref{table:category} presents a detailed breakdown of performance across the 13 categories of MM-SafetyBench. The data clearly shows that our ASO enhancement (+ Ours) is not just an artifact of average performance, but provides a consistent and general-purpose boost across nearly all categories and for all base attacks. This effect is evident at both extremes. First, for "hard" categories where base attacks like QR Attack and SI Attack have a very low ASR (e.g., Fraud, HateSpeech, Physical Harm), our ASO method still finds and optimizes a vulnerability signal, often resulting in a significant ASR increase, nearly doubling the success rate in the case of Fraud. Second, at the other extreme, for an already potent attack like HIMRD which achieves near-saturation ASR (>95\%) on many categories, our ASO framework still extracts additional value. While the ASR gain is marginal in these saturated cases (e.g., EconomicHarm from 95.1\% to 97.5\%), the Harmfulness Score (HS) consistently and significantly increases (e.g., from 9.6 to 10.1). This demonstrates that our RL-optimized style makes the successful jailbreak response more confident and semantically harmful. In summary, this fine-grained analysis validates that ASO is a robust and universal amplifier, capable of both turning weak vulnerability signals into viable attacks and hardening existing attacks.

\subsection{Ablation Study}

To quantify the distinct contributions of our two-stage methodology, we performed a critical ablation study, with results shown in Table~\ref{table:ablation}. This table dissects ASO's performance by comparing the Original base attack ASR against (1) + Probing, which applies only the naive, non-optimized best style ($S^*$) found in Section~\ref{sec:style-probing}, and (2) ++ Enhance, which represents our full RL-optimized ASO framework. The results clearly demonstrate a two-stage improvement. The + Probing step consistently provides a minor but positive ASR boost over the Original baseline across all models and attacks (e.g., SI Attack on Qwen3-VL improves from 39.31\% to 40.62\%), validating that stylistic sensitivity is a real and exploitable vulnerability. However, the data confirms that the vast majority of the performance gain is attributable to our RL-based enhancement stage. In all test cases, the ++ Enhance step provides a much more significant ASR jump over the + Probing step (e.g., SI Attack on LLaVA-OV-1.5 leaps from 39.96\% to 44.25\%). This strongly validates our core hypothesis: while identifying a sensitive style (Probing) is useful, it is the adversarial optimization (Enhance) via our algorithm that unlocks the style's full potential and creates a truly potent attack.

\begin{table}[t]
	\centering  
\begin{tabular}{l|c|c}
\toprule
\multicolumn{1}{c|}{\multirow{2}[0]{*}{\textbf{Method}}} & \textbf{Qwen3-VL} & \textbf{LLaVA-OV-1.5} \\
      & \textbf{ASR}   & \textbf{ASR} \\
\midrule
QR Attack \cite{liu2024mm} & 38.99\% & 37.80\% \\
\textbf{ + Probing} & 40.48\% & 40.12\% \\
\textbf{ ++ Enhance} & \textbf{42.98\%} & \textbf{44.35\%} \\
\midrule
SI Attack \cite{zhao2025jailbreaking} & 39.31\% & 37.82\% \\
\textbf{ + Probing} & 40.62\% & 39.96\% \\
\textbf{ ++ Enhance} & \textbf{42.58\%} & \textbf{44.25\%} \\
\midrule
HIMRD \cite{ma2025heuristic} & 87.38\% & 52.92\% \\
\textbf{ + Probing} & 89.17\% & 54.23\% \\
\textbf{ ++ Enhance} & \textbf{89.52\%} & \textbf{55.42\%} \\
\bottomrule
\end{tabular}%

	\caption{Ablation study of different components.}  
	\label{table:ablation}  
\end{table}

\section{Conclusion}
In this work, we systematically investigated innate stylistic sensitivity—a novel, non-content-based VLM vulnerability unexploited by content-based jailbreaks. We leverage this by introducing ASO, a general-purpose, plug-and-play enhancement framework to amplify existing SOTA attacks. Our optimization, enabled by an ODE-to-SDE conversion for flow models and guided by a dense Structurally-Tiered Reward Function, efficiently discovers potent hybrid triggers. Experiments on SOTA VLMs and SOTA attacks conclusively show ASO consistently amplifies the Attack Success Rate. Our findings prove VLM safety is a function of how (style) not just what (content), revealing a larger attack surface and the need for defenses beyond content-centric filters.
{
    \small
    \bibliographystyle{ieeenat_fullname}
    \bibliography{main}
}


\end{document}